\documentclass[conference]{IEEEtran}
\IEEEoverridecommandlockouts

\usepackage{graphics} 
\usepackage{epsfig} 
\usepackage{mathptmx} 
\usepackage{balance}

\usepackage{multirow}
\usepackage{hhline}

\usepackage{times} 
\usepackage{amsmath} 
\usepackage{amssymb}  

\usepackage[utf8]{inputenc}
\usepackage{algorithmicx}
\usepackage[ruled]{algorithm}
\usepackage{algpseudocode}

\usepackage{threeparttable}

\usepackage{colortbl}  
\usepackage{xcolor}
\usepackage{array}

\usepackage{cite}
\usepackage{graphics}
\usepackage{amsmath,amssymb,amsfonts}
\usepackage{epsfig}
\usepackage{mathptmx}

\usepackage{graphicx}
\usepackage{textcomp}
\usepackage{xcolor}
\def\BibTeX{{\rm B\kern-.05em{\sc i\kern-.025em b}\kern-.08em
    T\kern-.1667em\lower.7ex\hbox{E}\kern-.125emX}}
\begin{document}

\title{TADP: Task-Aware Deformable Prediction for Single-Stage 3D Object Detection \\

\thanks{This work is supported by Key Research and Development Plan of Shaanxi Province (China) under grant no. 2022GY-080.}
}

\author{\IEEEauthorblockN{1\textsuperscript{st} Su Wang}
\IEEEauthorblockA{\textit{School of Software Engineering} \\
\textit{Xi'an Jiaotong University}\\
Xi'an, China \\
wangsu@stu.xjtu.edu.cn}
\and
\IEEEauthorblockN{2\textsuperscript{nd} Yaochen Li*}
\IEEEauthorblockA{\textit{School of Software Engineering} \\
\textit{Xi'an Jiaotong University}\\
Xi'an, China \\
yaochenli@mail.xjtu.edu.cn}
\and
\IEEEauthorblockN{3\textsuperscript{rd} Min Yang}
\IEEEauthorblockA{\textit{School of Software Engineering} \\
\textit{Xi'an Jiaotong University}\\
Xi'an, China \\
yangmin3056@stu.xjtu.edu.cn}
\and
\IEEEauthorblockN{4\textsuperscript{th} Jiaohao Nie}
\IEEEauthorblockA{\textit{School of Electrical and Electronic Engineering} \\
\textit{Nanyang Technological University}\\
Singapore \\
jnie002@e.ntu.edu.sg}
\and
\IEEEauthorblockN{5\textsuperscript{th} Chang Liu}
\IEEEauthorblockA{
\textit{CSSC Systems Engineering}\\
\textit{Research Institute}\\
Beijing, China \\
liuc1100101110@163.com}
\and
\IEEEauthorblockN{6\textsuperscript{th} Yuehu Liu}
\IEEEauthorblockA{\textit{College of Artificial Intelligence} \\
\textit{Xi'an Jiaotong University}\\
Xi'an, China \\
liuyh@mail.xjtu.edu.cn}
}

\maketitle

\begin{abstract}
Most single-stage 3D object detectors complete different tasks with the same extracted features. Nevertheless, it is impossible to project features into a common space that is adaptive for all the tasks. We present a novel task-aware deformable prediction (TADP) method for single-stage 3D object detection to solve this problem. Firstly, a triple feature refinement aggregation module is designed to extract three-level features adaptively. Additionally, we design the multi-scale feature aggregation block to fuse multi-scale features in a scale-aware manner. Finally, the prediction of each task is deformed with the designed plug-and-play task-aware deformation head. It can percept the emphasis and interaction of each task. We also designed three different deformation modules. The experimental results demonstrate that the proposed deformation head shows good results on other detection methods. The experimental results on the KITTI dataset demonstrate that the car mAP is 80.91\%, surpassing many state-of-the-art methods on the KITTI benchmark.

\end{abstract}

\begin{IEEEkeywords}
Point Cloud, Single-stage Detection, 3D Object Detection, Task-aware
\end{IEEEkeywords}

\section{Introduction}
 With the development of automatic driving, lidar sensors are becoming more and more critical in 3D object detection. In general, object detection is divided into two categories in point clouds: single-stage and two-stage. Compared with single-stage methods, two-stage methods have higher accuracy but have more computational costs. Due to the limited power of the embedded hardware for automatic driving cars and robots. We improve the accuracy of the single-stage detector to achieve low computational cost and high accuracy simultaneously. However, there are a large number of sparse and disordered points that pose a significant challenge to the detectors. Some two-stage methods are inspired by PointNet\cite{c1} and PointNet++\cite{c2}, which through point-based methods to extract scene features. Some single-stage methods are inspired by VoxelNet\cite{c3}. These methods encode point clouds orderly, such as voxels and pillars\cite{c4}.

\begin{figure}[htbp]
\centering{
\includegraphics[width=0.46\textwidth]{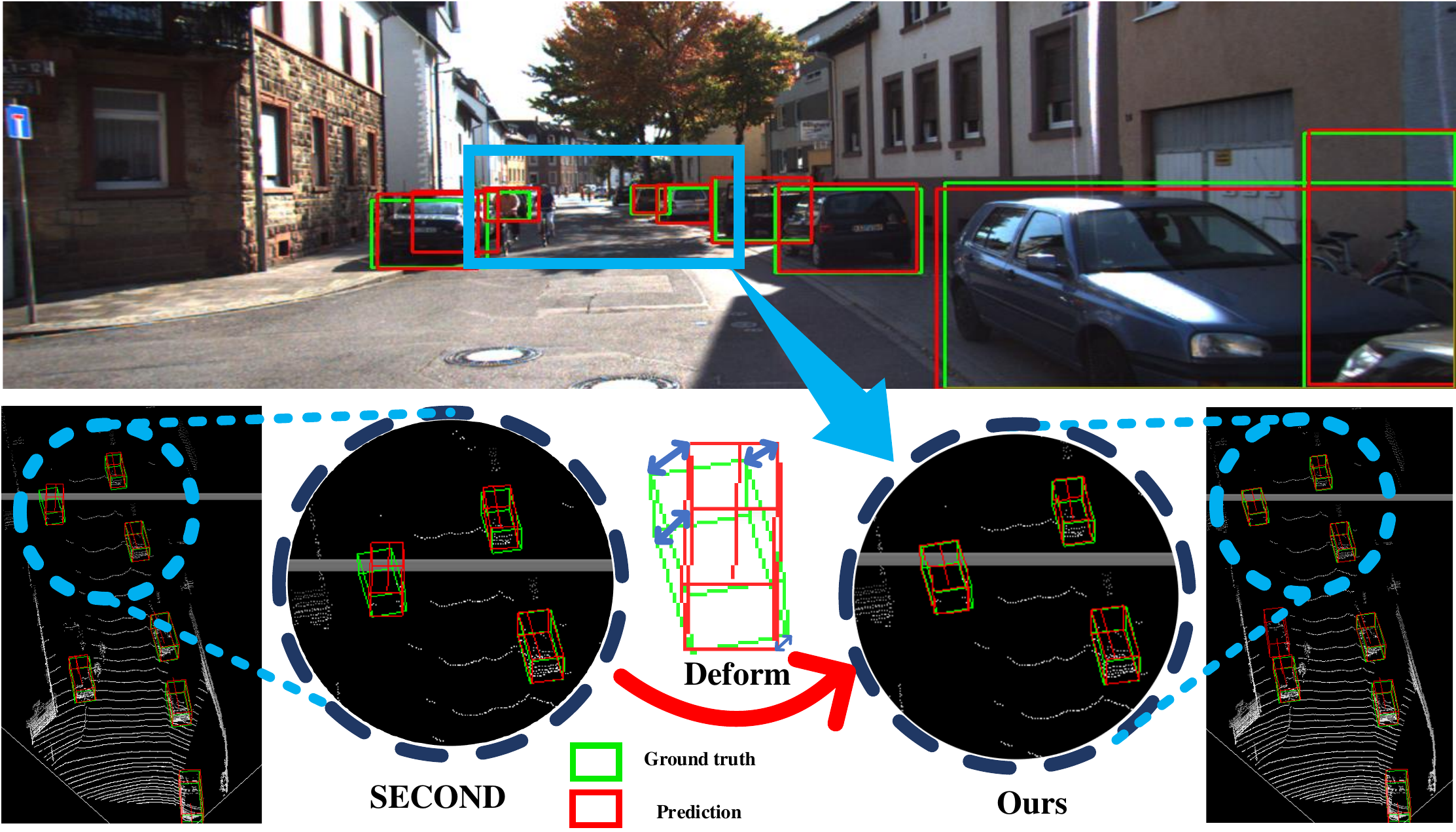}
}
\caption{Visualization of detection in street scenes. It shows the different detection results of SECOND\cite{c5} and our TADP. We use red arrows to indicate the biased optimization of our method compared to SECOND detection.}\label{fig1}
\end{figure}

Previous single-stage methods mainly focus on feature extraction networks instead of the effectiveness of the detection head. The features extracted in the single-stage detector are less accurate than those in the two-stage. Therefore, it is difficult to predict results precisely and may cause some deviation in the difference among tasks in single-stage detectors. Thus, the detection head of tasks should correspond with the task's feature. Therefore, a suitable detection head that can align tasks is crucial for single-stage detectors. 


To solve the shortcoming of existing methods, we design a triple feature refinement and aggregation module (TFRA), which refines features in three scales: the semantic, structural, and geometric scales, respectively. In previous methods, directly changing features into different scales and fusing them will cause information loss. Consequently, we design a multi-scale feature aggregation module (MSFA) for feature fusion. Due to the limited accuracy of the single-stage detection head. We propose a task-aware deformable head (TADH). The task perception stack can perceive each task's features and predict the semantic deformation map (DMap). Moreover, we use height attention to improve the sensitivity of the DMap, which can avoid the limitations of bird-eye view (BEV). Then we modify the prediction results of each task with different suitable deform strategies. Our proposed method dramatically improves the accuracy of the single-stage methods. It is worth mentioning that TADH is detachable and plug-and-play for other 3D detectors, and the experiments show that the accuracy has been improved when TADH is applied to other single-stage detectors. Our primary contribution is manifold.

\begin{itemize}
    \item We present an efficient and high-precision end-to-end single-stage detection network task-aware called TADP.
    \item We design three-level feature extraction and scale-aware fusion network to extract and fuse multi-scale features of point cloud scenes effectively.
    \item We propose a plug-and-play task-aware deformable head. It is adopted to optimize the prediction results of the task. Applying it to other detectors can also significantly improve accuracy.
\end{itemize}
Our method has achieved high performance and superior inference speed in 3D road scene detection in the KITTI\cite{c7} dataset.

\section{related work}

Two kinds of Lidar-based 3D object detectors. (\romannumeral1) The two-stage detectors generate RoI in the first stage and refine RoI in the second stage. (\romannumeral2) The single-stage object detectors directly generate regression classification and boundary boxes from the first stage. The two-stage method has the advantage of high accuracy. Despite the high accuracy of two-stage, single-stage methods are widely used due to their simpler structure and higher speed. With the development of single-stage methods, the accuracy of the two-stage can be gradually reached. They are becoming an important method for 3D perception in autonomous driving.

PointRCNN\cite{c8} is a two-stage detection method based on PointNet++, and the author proposes a point-based AnchorFree strategy. The first stage proposes regional suggestions, and the second stage refines the interior point features to adjust the 3D frame. PointFormer\cite{c9} uses three transformers to extract scene features and refine them. But it has a large computational cost. Voxel R-CNN\cite{c10} uses voxel-based rather than point-based methods and adopts voxel RoI pooling to extract proposal features in more detailedly. Votr\cite{c11} extracts contextual information between voxels by designing a voxel-based sparse transformer. SST\cite{c12} successfully increases the receptive field through the transformer, increasing the accuracy of small objects. However, the two-stage methods have the problems of over-computation and high computational cost.

The single-stage methods include VoxelNet, which first preprocesses disordered point clouds into regular voxels. Point-Pillar and SECOND segment point clouds into regular voxels and utilize sparse convolution and submanifold convolution to process segmented voxels. TANet\cite{c13} designs a triple attention module based on points to extract robust features. 3DSSD\cite{c15} devises a more advanced point-based fusion adoption strategy. CIA-SSD\cite{c16} proposes a voxel-based IOU-predictive perceptual detection head. However, the features of the single-stage are fragile and the results are unstable.

The proposed task-aware deformation for the detection head can be used to correct the misaligned prediction results. We aim to increase the prediction accuracy of the single-stage method with the controllable computational cost. Enables single-stage detection to be high-speed and high-precision at the same time.

\section{Method}
\begin{figure*}[htb]
	\centering
	\includegraphics[height=70mm, width=\textwidth]{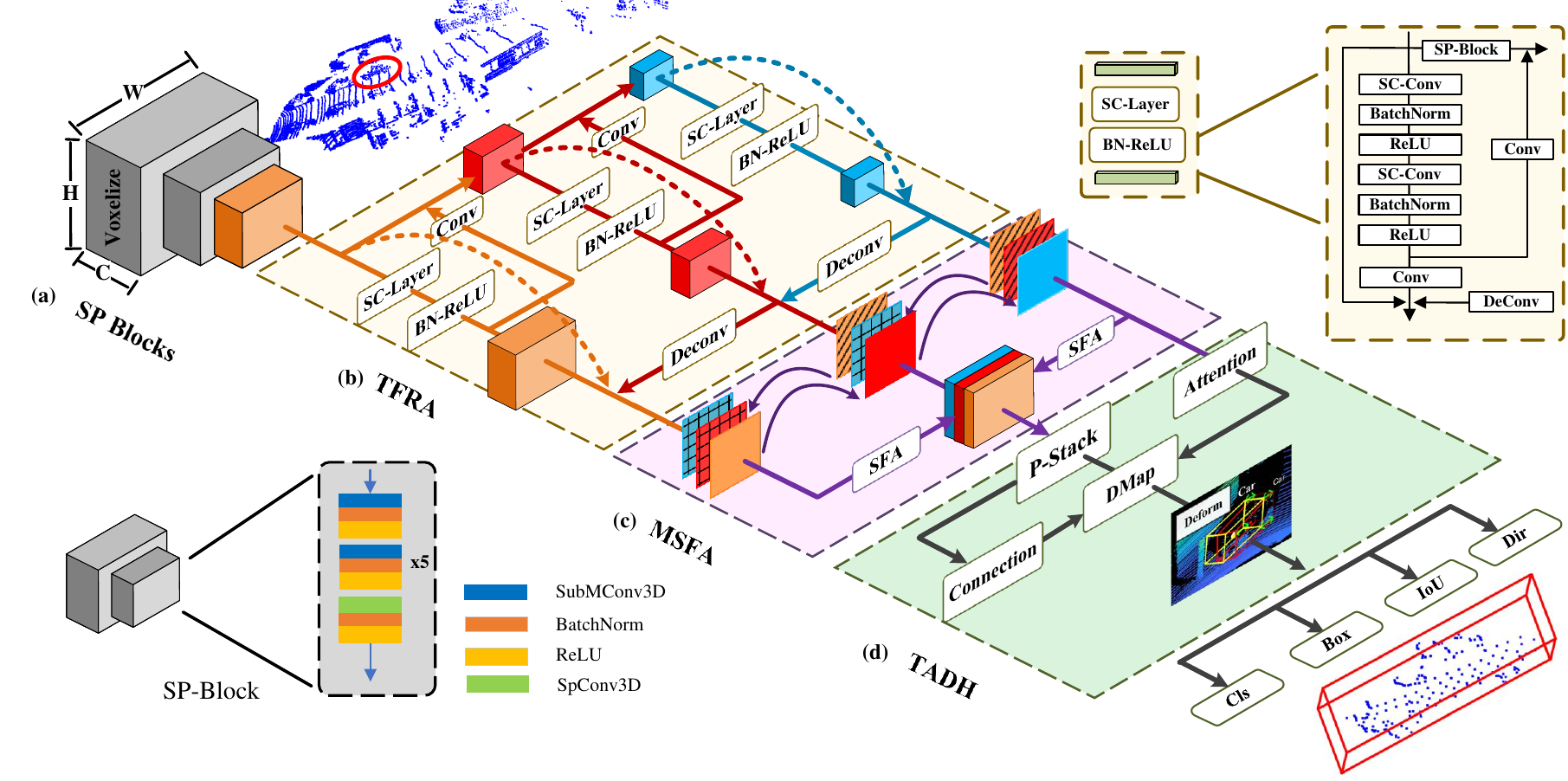}
        
        \vspace{-0.4cm}
	\caption{The figure shows the pipeline of our method TADP. (a) downsample and encode the point clouds. (b) show TFRA structure refine the three-scales features. (c) show the MSFA structure, scale maps Features and fused with SFA. (d) show the TADH, predict the deformation map, and deform the task prediction.}
	\label{fig2}
\vspace{-0.6cm}
\end{figure*}

We design a single-stage object detection method called task-aware deformable prediction for single-stage 3D object detection (TADP). Fig.~\ref{fig2} shows our designed method, which consists of three parts: (\romannumeral1)sparse  blocks (SP Blocks);
(\romannumeral2)triple features refine and fusion, which contains tripe feature refinement aggregation (TFRA) and multi-scale feature aggregation (MSFA); (\romannumeral3)task-aware deformable head (TADH).

\vspace{0cm}

\subsection{Point Cloud Feature Encoding} 

As shown in Fig.~\ref{fig2}, the Sparse Blocks (SP Blocks) encode the point clouds. We convert point clouds to voxel format like SECOND. We divide the scene into 40x1600x1400 voxels. The Sp Blocks' details are shown in Fig.~\ref{fig2}. To ensure the sparsity of sampled voxels, the Sp Blocks include submanifold sparse convolution\cite{c19} and sparse convolution\cite{c18} used alternately. We use the bird-eye view (BEV) method to compress the features and the three features after the SP Blocks are used as the input of the next module.

\subsection{Triple Feature Refine Aggregation}

This part mainly describes the TFRA module. The three-level structure is used for feature multi-scale refinement, mainly divided into two modules, a three-branch feature refinement module and a Multi-Scale Feature Aggregation module.

\subsubsection{Triple Feature Refine and fusion} ${\rm{     }}$

We are inspired by the feature pyramid. We designed the TFRA module to extract scene information on various scales. We divide the network into three independent branches, respectively extracting semantic, structural and geometric scale features. Details are shown in Fig. 2(b). We refine the features by using the self-correcting layer. SC-Layer is a stack containing two SCConv\cite{c21} and a full connected layer, which is able to extract local and global features flexibly and has a variable receptive field. We use deconv to change the feature size, and design the self-residual, upward-residual, and downward-residual connection. The specific process is shown in Fig.~\ref{fig2}.



\subsubsection{Multi-scale feature aggregation}

 Directly fuse features of different sizes will cause information loss of fragile features. Aiming at this problem, we designed a feature fusion method called Muti-Scale Feature Aggregation (MSFA), as shown in Fig. 2(c). We devised the scale mapping (SM) method to map all other features to one feature. Different scales use different fusion methods to enhance different details. SM function is shown in Eq.\ref{eq2}:

\begin{equation}\label{eq2}
  \begin{split}
M^{x_m, x_n}&=P(x_n)\cdot B(C(x_m)) \\
F_M^{x_m, x_n}&=B(C(M^{x_m, x_n}+X_m)) 
  \end{split}
  \end{equation}

\noindent
where $M^{x_m,x_n}$ is the feature mapped from $x_n$ to $x_m$. $C(\cdot)$ stands for Conv. $B(\cdot)$ stands for BatchNorm, $P(\cdot)$ stands for average pooling. $F_SM^{x_m,x_n}$ represents the feature after scale mapping. We scale map each level feature with other features. As shown in Fig.~\ref{fig2}, the color represents the feature mapped with, and the pattern represents the meaning of the mapping. SFA uses softmax to establish feature dependencies for adaptive fusion.

\subsection{Task-Aware Defomable Head}

\begin{figure}[htbp]
\centering{
\includegraphics[height=30mm,width=0.45\textwidth]{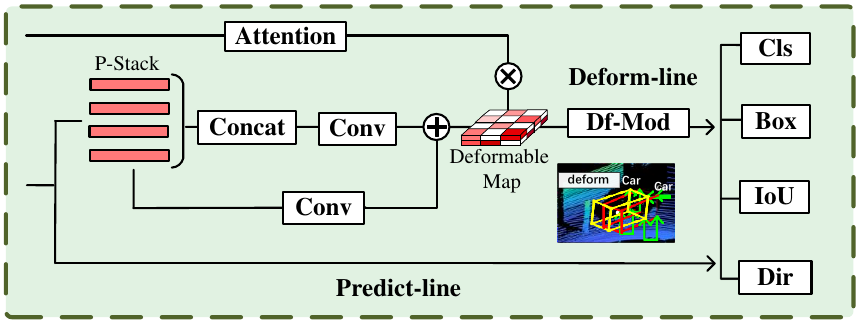}
}
\caption{The pipeline of TADH. The first branch of the features directly generates traditional task predictions through predict line. The second branch through P-Stack generates DMap. The upper features provide height attention. Finally, the deformation to the prediction result is generated through the Deform-line.}\label{fig3}
\vspace{-0.1cm}
\end{figure}

The previous detection head can not effectively distinguish the differences between each task, and the predicted features can not precisely match tasks. To solve the problems faced by the single-stage method, we designed a detection head module named Task-Aware Deformable Detection Head (TADH). Firstly, we introduce a task perception stack to sense each task. Then we generate a semantic deformation map (DMap) and introduce extra height attention to sensitize the DMap. Finally, we apply deformation in task prediction. The specific structure is shown in Fig.~\ref{fig3}.

\subsubsection{Task Perceptual Stack}

We design a task perceptual stack (P-Stack) to sense task focus and correct misalignment between tasks. The P-Stack contains multiple convolution layers, giving enough mutual receptive fields between tasks. $X_0^{perc}$ represents the refined feature. The P-Stack is composed of N consecutive fully connected layers with activation functions to compute aligned interaction stacks for different features. The perceptual stack formula is shown in Eq.\ref{eq3}.

\vspace{-0.5cm}
\begin{equation}\label{eq3}
    \begin{split}
        X_k^{perc}=B(\sigma(C_k(X_{k-1}^{prec})))
    \end{split}
\end{equation}
where $B,\sigma$ represents BatchNorm and ReLU function. $C_k$ represents the $k_{th}$ layer task interactive convolution. $X_k^{perc}$ represents the features stored in the $k_{th}$ perception stack. This stacking feature can well perceive the state of each task by adjusting the dislocation between tasks on the stack. It provides a task-aware pool for the next deformation maps.

\subsubsection{Deformation Map and Predict Deforme}

\begin{figure}[htbp]
\centering{
\includegraphics[height=40mm,width=0.45\textwidth]{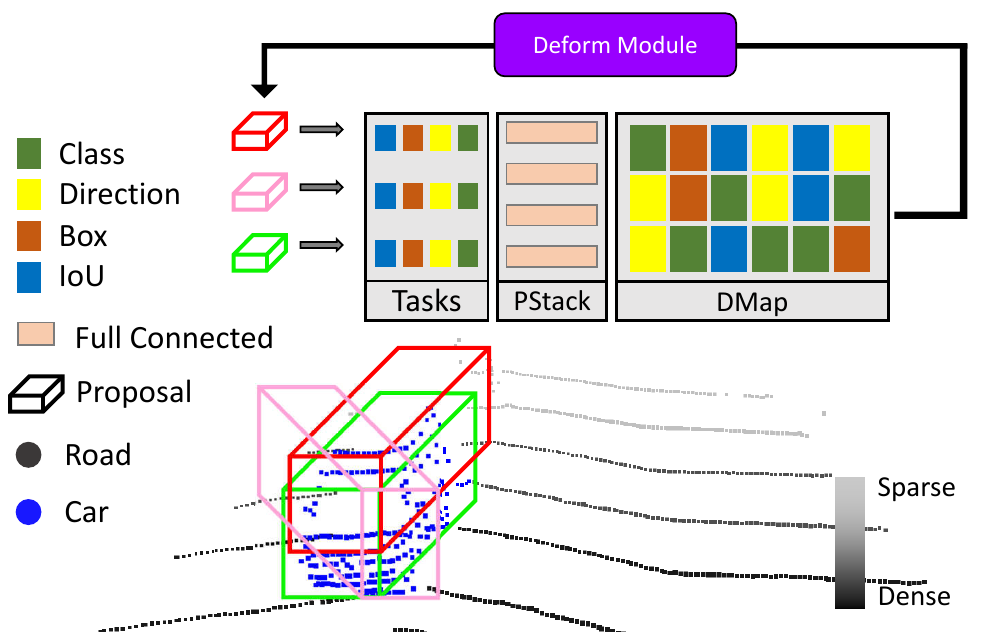}
}
\caption{The process and principle of PStack and DMap generating each task's deformation. Below shows the raw point cloud and proposals. Above shows the process of task alignment }\label{fig_new}

\end{figure}

In order to systematically optimize the prediction results, we design a semantic deformation map (DMap) so that each task learns its own deformation according to the DMap. The method of predicting DMap: First, concat the features of each layer in the P-Stack to generate $X^{perc}$. Generate DMap using $X^{perc}_N$ residual and $X^{perc}$. Using the height information of the geometric features to enhance the semantic information of the DMap. The process is shown in Fig.~\ref{fig_new}. It makes the DMap highly sensitive to compensate for the lack of BEV information.


\begin{figure}[htbp]
\centering{
\includegraphics[width=0.28\textwidth]{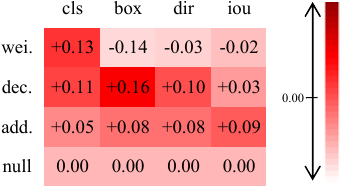}
}
\caption{Comparative experiments between different tasks and different deform modules. $Wei.$, $dec.$, $add.$, and $null$. We show how much the accuracy increases as a color chart. Based on the `$null$' color. }\label{fig4}
\end{figure}

We design three modules to deform different tasks. The first is the weight module, which generates deformation weights to deform prediction results. The second is the convolution module, which uses deformconv\cite{c22} to deform the prediction results. The third is an additional module that adds the deformation to the results. For four tasks: class, bounding box, direction, and IoU, we experiment on different tasks with different deform modules. The results of the experiment are shown in Fig.~\ref{fig4}. According to the experimental results, we choose the weight module for the classification task, the convolution module for the box and direction task, additional module for the IoU task. The algorithm for making prediction deformation is shown in Algorithm~\ref{Alg1}.

This detection head is plug-and-play, regardless of the height attention attached to the DMap. It is suitable for other single-stage point cloud object detection. TADH can significantly improve the accuracy of other methods.


\begin{minipage}{8cm}
\begin{algorithm}[H]
\caption{Tasks Deform Module}\label{Alg1}
\begin{algorithmic}[0.9]
\footnotesize
\Statex \textbf{Note:}
\Statex \quad \quad $T\in\{box,cls,dir,iou\}$
\Statex \quad \quad $P_T$ is the traditional prediction result.
\Statex \quad \quad $FC_T$ is task fully connected layer.
\Statex \quad \quad $DP_T$ is deformed task prediction.
\Statex \quad \quad $DefConv$ is deformconvlution.
\Procedure{Task Predict Deform}{$T,P_T,DM$} 
   \While{$t\in T$}
            \If{$t=cls$}
                \State $D_{cls}=FC_{cls}(DM)$
                \State $DP_{cls}=\sigma(\sqrt{P_{cls}\ast D_{cls}})$
            \EndIf
            \If{$t=box or t=dir$}
                \State $D_{box}=FC_{box}(DM)$
                \State $D_{dir}=FC_{dir}(DM)$
                \State $DP_{box}=DefConv(P_{box},D_{box})$
                \State $DP_{dir}=DefConv(P_{dir},D_{box})$
            \EndIf
            \If{$t=iou$}
                \State $D_{iou}=FC_{iou}(DM)$
                \State $DP_{iou}=\frac{D_{iou}+P_{iou}}{2}$  
            \EndIf
    
    \EndWhile
    
\State \Return $DP_t$
\EndProcedure
\end{algorithmic}
\end{algorithm}
\end{minipage}

\subsection{Loss Function}

We followed the general setup of loss functions in CIASSD\cite{c16} networks. Specifically, we use Focal loss for bounding box classification loss, Smooth-L1 loss for bounding box regression loss, and cross-entropy loss for orientation classification loss. Use SmoothL1 loss for IOU loss. Inspired by gIoU\cite{c23}, the traditional IoU fails to get correct regression when the anchor and ground truth are at certain angles. We set $\lambda=1, \mu=1, \omega=2, \delta=0.2$. $SML_1$ represent Smooth-L1 loss, and total loss $L_{total}$ is as Eq.\ref{eq 4}:

\vspace{-0.1cm}
\begin{equation}\label{eq 4}
  \begin{split}
  L_{giou}&=SML_1(1-giou)    \\
  L_{total}&=\lambda L_{cls}+\mu L_{iou}+\omega L_{box}+\delta L_{dri}
  \end{split}
  \end{equation}

\begin{figure*}[htb]
	\centering
	\includegraphics[width=\textwidth,height=70mm]{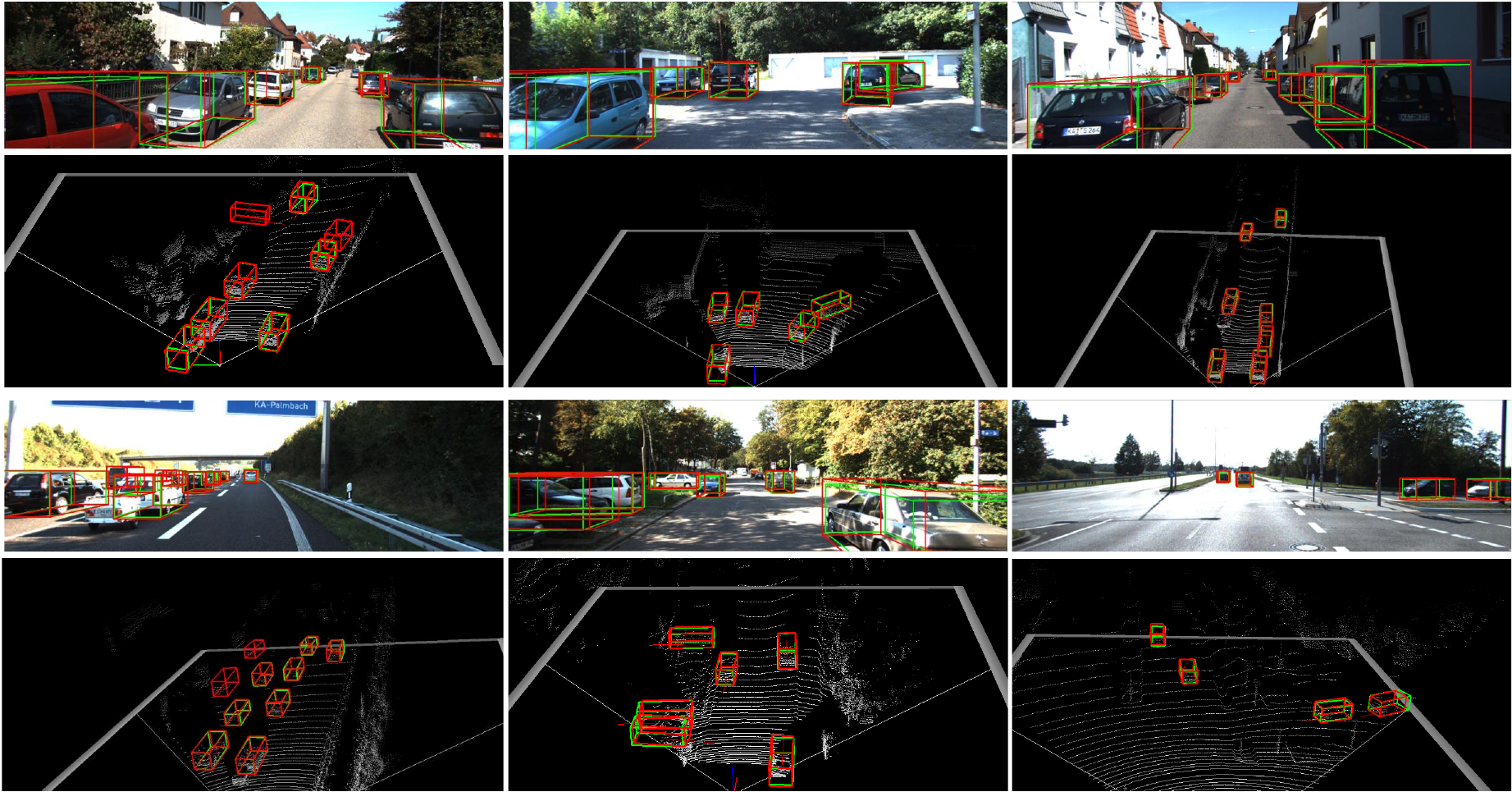}
	\caption{Our 3D detection results on the KITTI validation set are visualized. The ground truth is the green box, and the prediction box is the red box. The 3D detection box is projected to an RGB image.}
	\label{fig5}
\vspace{-0.7cm}
\end{figure*}

\section{Experiment}

The experiments use the KITTI dataset. We mainly detect car class. The KITTI dataset divides the model evaluation difficulty into Easy, Moderate, and Hard levels. The comparison experiment uses the test set and submits it on the KITTI benchmark. Ablation experiments are evaluated using the validation set because of restricted access to the test set. Fig.~\ref{fig5} shows the visualization of the detection results. According to the official KITTI evaluation metric, the 3D and BEV detection results are evaluated with mean average precision (mAP).

\subsection{Implementation Details}
To achieve efficient detection, we adopt data augmentation to eliminate similar classes. We filter out objects in difficulty levels that belong to something other than easy, moderate, and hard to improve the quality of positive samples. Then, taking similar classes of objects (such as van for car) as mitigation, the target of model confusion during training. The selected voxel sizes are [0.05m, 0.05m, 0.1m]. Therefore, the resulting voxel grid size is$1408\times1600\times40$.

The SC-layer in TFRA selects the $k3$ mode and applies the linear interpolation method for up-sampling. Using a $3\times3$ convolution kernel and a $5\times5$ convolution kernel to expand the receptive field. Then, a $1\times1$ convolution layer is used, and the size of each level is unified with deconvolution for triple fusion. P-Stack in TADH we set N=4.  We examine the influence of parameter changes of $N$ on the results through comparative experiments shown in Tab.~\ref{tab4}. It can be seen that the accuracy keeps increasing with the increase of $N$, but the slope gradually decreases. Considering the accuracy and computational cost, we choose $N=4$. We set the batch size to 4, trained on RTX3090 GPU, and set the epoch to 60. The Adam optimizer is used, the initial learning rate is set to 0.003, and the exponential decay factor is 0.4, which decays every ten cycles.

\begin{table}\scriptsize
\centering
\caption{AP11-based comparation with SOTA on KITTI dataset benchmark for car class detection} \label{tab1}
\resizebox{0.5\textwidth}{!}{\begin{tabular}{c|c|c|ccc} 
\hline\hline
\multirow{2}{*}{Type}     & \multirow{2}{*}{Method} & \multirow{2}{*}{Sens} & \multicolumn{3}{c}{$AP_{3D}(\%)$}  \\
                          &                         &                       & Easy & Mod & Hard     \\ 
\hline\hline
\multirow{16}{*}{2-stage} 
                          & AVOD(2018)\cite{c28}             & LIDAR+RGB             & 83.07  & 71.76 & 65.73        \\
                          & PI-RCNN(2020)\cite{c29}          & LIDAR+RGB             & 84.37  & 74.82 & 70.03        \\
                          & PointRCNN(2019)        & LIDAR                 & 86.96  & 75.64 & 70.70         \\
                          & F-ConvNet(2019)\cite{c30}       & LIDAR+RGB             & 87.36  & 76.39 & 66.69        \\
                          & UberATG-MMF(2019)       & LIDAR+RGB             & 88.40   & 77.43 & 70.22        \\
                          & Part-A2(2020)\cite{c33}           & LIDAR                 & 87.81  & 78.49 & 73.51        \\
                          & Pointformer(2021)       & LIDAR                 & 87.13  & 77.06 & 69.25        \\
                          & 3D-CVF(2020)\cite{c35}            & LIDAR+RGB             & 88.84  & 79.72 & 72.80         \\
            
                          & Sem-Aug(2022)\cite{c43}        & LIDAR+RGB                     & 86.69  & 78.06 & 73.85         \\
                          & Fast-CLOCs(2022)\cite{c37}        & LIDAR+RGB             & \textit{89.10}  & \textit{80.35} & \textit{76.99}         \\                 
\hline\hline
\multirow{13}{*}{1-stage} & VoxelNet(2018)          & LIDAR                 & 79.62  & 65.97 & 59.71        \\
                          & ContFuse(2018)\cite{c39}          & LIDAR+RGB             & 83.68  & 68.78 & 61.67        \\
                          & SECOND(2018)            & LIDAR                 & 83.34  & 72.55 & 65.82        \\
                          & PointPillars(2019)     & LIDAR                 & 82.58  & 74.31 & 68.99        \\
                          & TANet(2020)             & LIDAR                 & 84.39  & 75.94 & 68.82        \\
                          & 3DSSD(2020)             & LIDAR                 & 88.36  & 79.57 & 74.05        \\
                          & SASSD(2020)\cite{c41}             & LIDAR                 & 88.75  & 79.79 & 74.16        \\
                          & HVPR(2021)\cite{c42}              & LIDAR                 & 86.38  & 78.22 & 73.84        \\
                          & MGAF(2021)\cite{c16}           & LIDAR                 & 88.16  & 79.58 & 72.39        \\
                          & ACDet(2022)            & LIDAR                 & 88.47  & 78.85 & 73.86         \\
                          & IA-SSD(2022)\cite{c17}            & LIDAR                 & 88.34  & \textbf{80.13} & 74.04         \\
                          & TADP(ours)              & LIDAR                 & \textbf{88.93}  & 79.65 & \textbf{74.17}        \\
\hline\hline
\end{tabular}}
\vspace{-0.5cm}
\end{table}

\subsection{ Compared with State-of-The-Art Methods}
We compare our algorithm with state-of-the-art 3D road scene detection algorithms. As shown in Tab.~\ref{tab1}, TADP is compared with the state-of-the-art object detection methods for 3D road scenes in the table. The best effect in the single-stage is shown in bold. It can be seen from the table that our method ranks first in the easy and hard levels of car detection, with 88.93\% and 74.17\%. Better than all demonstrated single-stage detectors in easy and moderate levels. TADP outperforms most of the two-stage object detectors in the table above, such as Pointformer and other state-of-the-art detectors. As seen from Tab.~\ref{tab5}, our method can outperform state-of-the-art two-stage detectors in both running speed and average precision.

\begin{table}
\centering
\caption{Ablation experiments of TFRA and MSFA and TADH in TADP on KITTI validation dataset.}\label{tab2}
\begin{threeparttable}
\begin{tabular}{cccc|ccc} 
\hline\hline
\multirow{2}{*}{$TFRA$} & \multirow{2}{*}{$MSFA$} & \multirow{2}{*}{$TADH$} & \multirow{2}{*}{$TADH^{\dagger}$} & \multicolumn{3}{c}{$AP_{3D}(\%)$}  \\ 
            &            &           &         & Easy  & Mod   & Hard   \\ 
\hline\hline
           &            &            &         & 87.52 & 77.21 & 74.36  \\
\checkmark &            &            &         & 88.27 & 78.41 & 75.89  \\
\checkmark&\checkmark &            &            & 89.26 & 79.37 & 76.76  \\
\checkmark&\checkmark & \checkmark &            & 89.71 & 79.95 & 77.31  \\
\checkmark&\checkmark &            & \checkmark & \textbf{90.03} & \textbf{80.62} & \textbf{78.64}         \\
\hline\hline
\end{tabular}
\begin{tablenotes}
        \footnotesize
        \item[$\dagger$]: means the TADH use GIoU
      \end{tablenotes}
    \end{threeparttable}
\vspace{-0.5cm}
\end{table}

\begin{table}\scriptsize
\centering
\caption{Runtime (in milliseconds) and mAP (in AP11) compared to recent state-of-the-art secondary detectors}\label{tab5}
\begin{tabular}{l|lllll} 
\hline\hline
         & PointRCNN & Part-A2 & MGAF-3DSSD & 3D-CVF & Ours   \\ 
\hline\hline
time(ms) & 643       & 80      & 80         & 85        & \textbf{40.53}  \\
mAP      & 77.67     & 79.7    & 80.51      & 80.79      & \textbf{80.91}  \\
\hline\hline
\end{tabular}
\vspace{-0.5cm}
\end{table}

\begin{table}\scriptsize

\centering
\caption{Comparitive test of convolution layer parameter N in Percetion Stack}\label{tab4}
\resizebox{0.5\textwidth}{!}{\begin{tabular}{c|ccccccc} 
\hline\hline
Stack\_num & 0   & 1    & 2    & 3    & \textbf{4}    & 5    & 6     \\
\hline\hline
 Improvement/$AP_{3D}(\%)$   &  78.41 & +0.11 & +0.23 & +0.38 & \textbf{+0.49} & +0.58 & +0.64  \\
\hline\hline
\end{tabular}}
\end{table}

\subsection{Ablation Experiments}

We conduct ablation experiments on the validation set of KITTI. Where $TADH^{\dagger}$ in Tab.~\ref{tab2} represents TADH with GIoU. The data in Tab.~\ref{tab2} shows that our designed TADH improves the overall network by 0.55\%, 0.78\%, and 0.76\% for easy, moderate, and hard levels, respectively. Moreover, we changed to gIoU for better results, with easy, moderate, and hard levels increasing by 0.32\%, 0.67\%, and 1.33\%, respectively. MSFA improved the easy, moderate and hard classes by 0.99\%, 0.96\%, and 0.87\%, respectively, in the experiment. It can be seen from the table that TADH can greatly optimize the detection results for moderate and hard classes. MSFA can better extract multi-scale features, up to 1\% improvement for easy class.

\begin{table}
\centering
\caption{Comparative test of car class detection after TADH insertion in some one-stage networks.}\label{tab3}
\begin{tabular}{c|ccc} 
\hline\hline
\multirow{2}{*}{Method} & \multicolumn{3}{c}{$AP_{3D}$(\%)}                \\
                        & Easy         & Mod          & Hard          \\ 
\hline\hline
SECOND                  & 83,52        & 73.63        & 67.21         \\
SECOND+DH               & \textbf{84.44}        & \textbf{74.25}   &\textbf{67.61}       \\
 Improvement                & +0.92         & +0.62         & +0.40          \\
\hline
VoxelNet                & 79.62        & 65.97        & 59.71         \\
VoxelNet+DH             & \textbf{80.78}        & \textbf{66.82}     & \textbf{60.35}         \\
 Improvement                & +1.16         & +0.85         & +0.64          \\
\hline
TANet                   & 85.42         & 76.34        & 69.92          \\
TANet+DH                & \textbf{86.40}        & \textbf{77.06}        & \textbf{70.39}    \\
 Improvement                & +0.98         & +0.72         & +0.47          \\
\hline\hline
\end{tabular}

\end{table}

\subsection{TADH Comparison Experiments}
In order to verify whether the TADH head can improve the detection effect of the single-stage detector, we spliced TADH into other backbones. We used SECOND, VoxelNet, and TANet networks for comparative experiments on the KITTI validation set. As can be seen from Tab.~\ref{tab3}, our TADH can significantly improve the detection effect of the backbone network in the one-stage method. This proves that our task deformation head can effectively correct the predicted misalignment results and be sensitive to capturing distant features of 3D objects.

\section{CONCLUSIONS}

This paper proposes a new point cloud single-stage task-aware deformable detector. To solve the situation that most methods of detection tasks are not aligned. Our main contribution includes triple-level extract and aggregate 3D features with multiple scales. It is worth noting that we propose a plug-and-play TADH, which predicts a sensitive deformation map to deform predicted results and reduce the misalignment of features in all tasks. Experiments show that our TADP achieves a really high accuracy performance on the KITTI benchmark. Moreover, the designed plug-and-play TADH can greatly improve the detection accuracy of existing single-stage detectors.


\end{document}